# Thing Foundational Ontology: ThingFO v1.3's Terms, Properties, Relationships and Axioms

**Luis Olsina**

GIDIS_Web, Engineering School, UNLPam, General Pico, LP, Argentina
`olsinal@ing.unlpam.edu.ar`

**Abstract.** This preprint specifies and defines all terms, properties, relationships and axioms of ThingFO (*Thing Foundational Ontology*) v1.3, which is a slightly updated version of its predecessor, ThingFO v1.2. It is an ontology for particular and universal Things placed at the foundational level in the context of a five-tier ontological architecture named FCD-OntoArch (*Foundational, Core, Domain and instance Ontological Architecture for sciences*). Figure 2 depicts its five tiers, which entail Foundational, Core, Top-Domain, Low-Domain and Instance levels. Two guidelines and three rules that guide the placement and constraints of ontologies in this ontological architecture are documented in a separate section. Each level is populated with ontological components or, in other words, ontologies. Ontologies at the same level can be related to each other, except at the foundational level, where only the ThingFO ontology is found. In addition, ontologies' terms and relationships at lower levels can be semantically enriched by ontologies' terms and relationships from the higher levels. ThingFO and ontologies at the core level such as ProcessCO, SituationCO, among others, are domain independent or neutral. ThingFO is made up of three main concepts, namely: Thing, Thing Category, and Assertion that represents human expressions about different aspects of particular and universal Things. Figure 1 shows the conceptualization of ThingFO specified in the UML language. Note that annotations of updates from the previous version (v1.2) to the current one (v1.3) can be found in Appendix A.

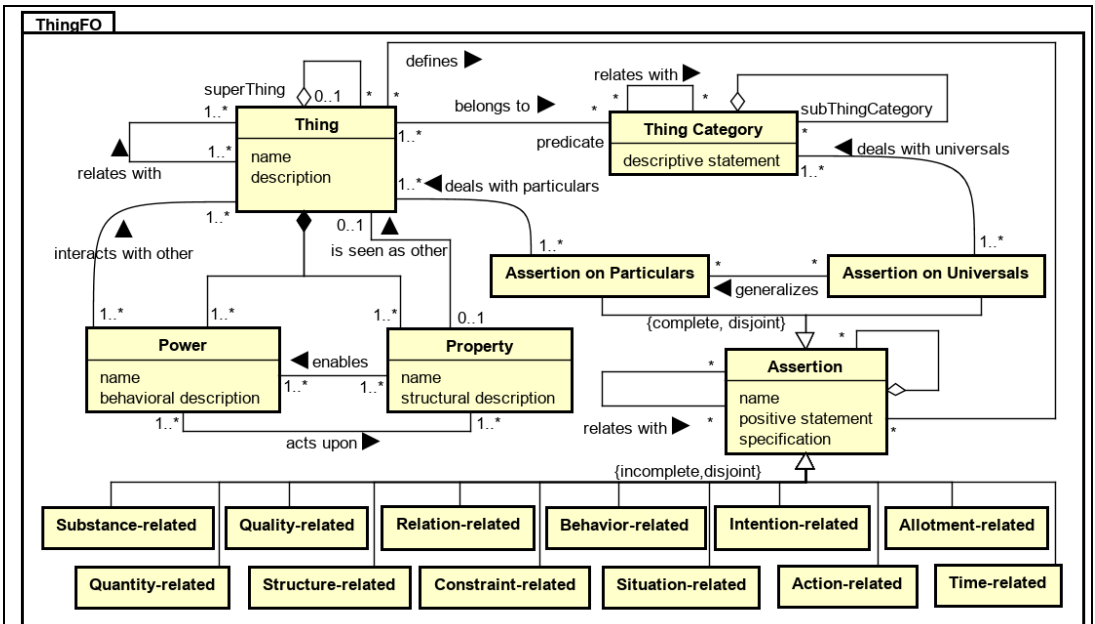

**Figure 1.** ThingFO v1.3: Foundational Ontology for Things, which is placed at the foundational level in the ontological architecture depicted in Figure 2. Annotations of updates from the previous version (v1.2) [11] to the current one (v1.3) can be found in Appendix A.

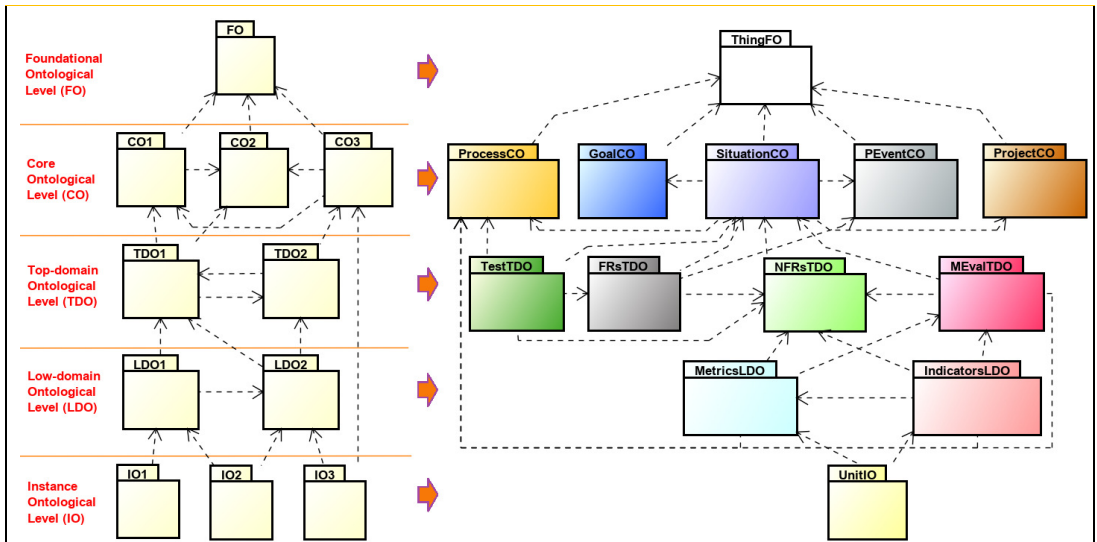

**Figure 2.** Allocating the ThingFO component or module in the context of the five-tier ontological architecture so-called FCD-OntoArch (*Foundational, Core, Domain and instance Ontological Architecture for sciences*) [5, 11]. ThingFO and ontologies at the core level such as SituationCO [10] and ProcessCO [2, 3] are domain independent or neutral.

## Thing Component – ThingFO v1.3's Terms

| Term | Definition |
|---|---|
| **Thing** (synonym: **Particular Thing**, **Object**, **Entity**, **Instance**, **Individual**) | Class or type of a perceivable or conceivable object, or its individuals of a given particular world. <br><br> Note 1: A Thing as a class represents and implies unique individuals or instances, not a universal category. Therefore, a particular Thing results in instances, whereas a universal Thing (i.e., a Thing Category) does not result in instances, at least with the valuable meaning of individual in a given particular world. <br><br> Note 2: A Thing is not a particular Thing without its Properties and its Powers, so "*things, properties and powers all emerge simultaneously to form a unity*" … "*Things, properties and powers are necessary and sufficient for the existence of this unity*" [1]. <br><br> Note 3: A Thing cannot exist or be in spatiotemporal isolation from other Things in a given particular world. In other words, a target Thing is always surrounded by other context Things, in any particular situation. <br><br> Note 4: In contrast to a particular Thing, that is, a particular class or subclass, the individuals of a class or subclass are not instantiated by further entities. <br><br> Note 5: A subclass of a particular Thing as a particular class can be represented at any lower level of the ontological architecture (depicted in Figure 2) except at the Instance Ontological Level, where only particular individuals are or exist. See Rule #3, in the "Guidelines and Rules" Section. |

| **Property** | It refers to the intrinsic constitution, structure, or parts of a particular Thing. |
| --- | --- |
| | Note 1: A Property is one member of the triad that conforms the unique identity named Thing. |
| | Note 2: A Property, which is one member of the triad that conforms a particular Thing can be seen as another particular Thing in another situation with its own Properties and Powers. |
| **Power** | It refers to what a particular Thing does, can do or behave. |
| | Note 1: A Power is one member of the triad that conforms the unique identity named Thing. |
| | Note 2: According to Fleetwood "*Powers are the way of acting of a things' properties; powers are a things' properties in action*" [1]. Also, he states that "*Things have properties, these properties instantiate […] acting powers, and this ensemble of things, properties and powers cause any events that might occur*". |
| **Thing Category** (synonym: **Entity Category**, **Universal Thing**) | Class or type that represents a category that predicates on particular Things conceived by a human being's mind for abstraction and classification purposes. |
| | Note 1: A Thing Category does not exist, is or can be in a given particular world as a Thing does. Conversely, it may only be formed or developed mentally by human beings. |
| | Note 2: A Thing Category as universal does not result in instances –at least with the valuable meaning of individual- but rather can be represented by more specific sub-categories of universal Things. |
| **Assertion** (synonym: **Human Expression**) | Class or type that represents a positive and explicit statement or expression that somebody makes about something concerning Things, or their categories, based on thoughts, perceptions, facts, intuitions, intentions and/or beliefs, conceived with an attempt to provide current or subsequent evidence. |
| | Note 1: The part of the previous definition that indicates "…*about something concerning Things…*" means, for example, about the substance, structure, behavior, relations, situations, quantity, quality, among other aspects of Things. |
| | Note 2: The part of the previous definition that indicates "…*statement or expression that somebody makes…*" means that a concrete human being –as a particular Thing- defines or conceives Assertions. |
| | Note 3: In order to be valuable, actionable and ultimately useful for any science, an Assertion should to a great extent be verified and validated by theoretical and/or empirical evidence. |
| | Note 4: An Assertion and its instances can be represented and modeled by means of informal, semiformal or formal expressions and specification languages. |
| | Note 5: ISO 21838-1 [4] defines the term "expression" as "*word or group of words or corresponding symbols that* |
| | *can be used in making an assertion*". |
| **Assertion on Particulars** | It is an Assertion that somebody makes about something of one or more particular Things. |
| **Assertion on Universals** | It is an Assertion that somebody makes about something of one or more Thing Categories. |
| **Action-related Assertion** | It is an Assertion related to the interaction and happening of Things since acting Powers cause any events that might occur.<br><br>Note 1: Particular Things can interact to each other, just as a Thing can act upon itself. See axioms A2 and A3.<br><br>Note 2: Interrelated Things interact to each other conforming particular situations, i.e., specific circumstances, episodes and events that are of interest for an intended agent.<br><br>Note 3: Interactions among Things both target entities and context entities in particular situations can be abstracted in generic situations. |
| **Allotment-related Assertion** | It is an Assertion related to the assignment of something, which implies the assignment of a Thing to itself or to other Things.<br><br>Note: For example, a particular resource (method, tool, person, etc.) is assigned to a task in a particular situation. Or, the specific amount of time a person gives him/herself to do an assignment. Or, the specific amount of time a professor gives their students to take a test. |
| **Behavior-related Assertion** | It is an Assertion related to the Power, which represents the capability and responsibility that a particular Thing has and/or exhibits.<br><br>Note: Behavior can be specified for particulars and can also be generalized for universals. |
| **Constraint-related Assertion** | It is an Assertion related to the specification of restrictions or conditions imposed on Things, Properties, relationships, interactions or Thing Categories that must be satisfied or evaluated to true in given situations or events.<br><br>Note: Constraint-related Assertions can be specified for both particulars and universals. |
| **Intention-related Assertion** | It is an Assertion related to the aim to be achieved by somebody.<br><br>Note 1: The statement of an Intention-related Assertion considers the propositional content of a goal purpose in a given situation and time frame.<br><br>Note 2: Intention-related Assertions can be specified for both particulars and universals. |
| **Quality-related Assertion** | It is an Assertion related to the requirements and constraints to be specified regarding the quality (distinguishing characteristic, attribute, or statement item) for a Thing and possibly related entities, which may be evaluable.<br><br>Note 1: Quality (cost, etc.) requirements and constraints can be specified for a particular Thing in terms of its Properties or Powers, or in terms of both as a whole. |
| | Note 2: Quality requirements and constraints can be specified for particulars and can also be abstracted or generalized for universals. |
| **Quantity-related Assertion** | It is an Assertion related to the countable, measurable and evaluable aspect of a Thing and possibly related entities, which can be specified by means of symbolic or numerical expressions. |
| | Note 1: Qualities of Things can be measured, evaluated and analyzed by specifying Quantity-related Assertions and strategies as resources. |
| | Note 2: A quantity or a relationship between quantities can be formalized, for instance, by mathematical, statistical or logical expressions. |
| | Note 3: Quantity-related Assertions can be specified for both particulars and universals. |
| **Relation-related Assertion** | It is an Assertion related to logical or natural associations between two or more Things and their categories. |
| | Note 1: A Thing cannot exist or be in spatiotemporal isolation from other Things in a given particular world. Therefore, a Thing is related to other Things. |
| | Note 2: Relationships can be specified for particular Things (between classes or between instances and classes, or between instances), and can also be represented for Thing Categories. |
| **Situation-related Assertion** | It is an Assertion related to the combination of circumstances, episodes, and relationships/events between target Things and context entities that surround them, or their categories, which is of interest or meaningful to be represented or modeled for an intended agent. |
| | Note 1: A Situation can be represented statically or dynamically depending on the intention of the agent. |
| | Note 2: Situations can be specified for particulars and can also be generalized for universals. |
| **Structure-related Assertion** | It is an Assertion related to the Property, which represents the intrinsic constitution, structure, or parts of a particular Thing. |
| | Note: Structural aspects can be specified for particulars and can also be abstracted for universals. |
| **Substance-related Assertion** | It is an Assertion related to the ontological significance and essential import of a Thing as a whole entity, or of a set of Things. |
| | Note: Substance aspects can be specified for particulars and can also be abstracted for universals. |
| **Time-related Assertion** | It is an Assertion related to the time as a Thing or its Properties or Power, which can imply specifying temporal boundaries or limits, among other aspects, for different situations and events. |
| | Note: Time aspects can be specified for particulars and can also be abstracted for universals. |

*Amount of Own Terms: 19*

### Thing Component – ThingFO v1.3's Properties or Attributes

| Term | Property | Definition |
|---|---|---|
| **Thing** | **name** | Label or name that identifies the particular Thing. |
| | **description** | An unambiguous textual statement describing a particular Thing. |
| **Property** | **name** | Label or name that identifies the Property of a Thing. |
| | **structural description** | An unambiguous textual statement describing the Property of a Thing in terms of its constituents, structure, or parts. |
| **Power** | **name** | Label or name that identifies the Power of a Thing. |
| | **behavioral description** | An unambiguous textual statement describing the Power of a Thing in terms of responsibilities, operations or actions. |
| **Thing Category** | **descriptive statement** | An unambiguous textual description of the category purpose as universal. |
| | | Note: The description of the category can be based on some Properties of particular Things, or some Powers of particular Things, or both. |
| **Assertion** | **name** | Label or name that identifies the Assertion. |
| | **positive statement** | An explicit declaration of the Assertion to be defined and expressed. |
| | | Note 1: Regarding a particular Thing or category, a positive statement refers to what it is, was, or will be, and contains no indication of approval or disapproval. |
| | | Note 2: A positive statement should be based on current or subsequent empirical evidence. |
| | **specification** | The explicit and detailed representation or model of the Assertion in a given language. |
| | | Note 1: Assertions can be modeled by means of informal, semiformal or formal expressions and specification languages. |
| | | Note 2: A specification can include text in natural language, mathematical and/or logical expressions, sketches, well-formed models and diagrams, multimedia resources, among other representations. |

*Amount of Properties: 10*

## Thing Component – ThingFO v1.3's Non-taxonomic Relationships

| Relationship | Definition |
|---|---|
| **acts upon** | A Power acts upon one or more Properties, so it can look at them or update the status of the Thing's properties.<br><br>Note: Note that this relationship represents internal actions, i.e., on the same Thing, not on other Things. This constraint is specified by axiom A2. |
| **belongs to** | Particulars Things may belong to none or more Thing Categories.<br><br>Note: In other words, a Thing Category predicates about a set of particular Things and their instances. |
| **deals with particulars** | An Assertion on Particulars deals with particular Things, both classes/subtypes and instances. |
| **deals with universals** | An Assertion on Universals deals with universal Things, which are Ccategories. |
| **defines** | A Thing defines none or many Assertions.<br><br>Note: For example, a particular Thing such as a Human Agent defines or conceives Assertions, such as Goals, Situations, among many others. |
| **enables** | A Property enables the Powers of a particular Thing.<br><br>Note 1: Because the Properties of a Thing are there, the Entity behavior can be enabled and manifested.<br><br>Note 2: Note that this relationship is restricted by axiom A1. |
| **generalizes** | An Assertion on Universals abstracts none or more Assertions on Particulars. |
| **interacts with other** | Due to the Power of a Thing, particular Things interact with each other.<br><br>Note: Note that this relationship represents actions on other Things, not on the same Thing. This constraint is specified by axiom A3. |
| **is seen as other** | A Property most of the time is seen as another Thing. |
| **relates with (x3)** | A Thing relates to other particular Things. |
| | A Thing Category may be related to other universal Things. |
| | An Assertion may be related to other Assertions. |

*Amount of non-taxonomic relationships: 12*

# Thing Component – ThingFO v1.3's Axioms

**A1 description**: *All Property of a Thing enables only its Powers.*
**A1 specification**:

$$1)\ \forall t, \forall prop, \forall pow: [Thing(t) \land Property(prop) \land partOf(prop, t) \\ \land Power(pow) \land enables(prop, pow) \rightarrow partOf(pow, t)]$$

**A2 description**: *The Power of a Thing only acts upon its Properties.*
**A2 specification**:

$$2)\ \forall t, \forall pow, \forall prop: [Thing(t) \land Power(pow) \land partOf(pow, t) \land Property(prop) \\ \land actsUpon(pow, prop) \rightarrow partOf(prop, t)]$$

**A3 description**: *The Power of a Thing only interacts with other Things.*
**A3 specification**:

$$3)\ \forall t, \forall pow: [Thing(t) \land Power(pow) \land partOf(pow, t) \\ \rightarrow \neg interactsWithOther(pow, t)]$$

# ThingFO and lower Ontologies in the context of FCD-OntoArch – Guidelines and Rules

Some guidelines and rules guide the incorporation and placement of ontologies at the FCD-OntoArch levels represented in Figure 2:

• Guideline #1: Any ontology conceptualization/formalization developed as an artifact cannot be conceived in isolation from an explicit specification of a layered ontological architecture. Therefore, a foundational ontology must be found at the upper or top level (Foundational Ontological Level in Figure 2, or FO level for short) of the ontological architecture.

• Guideline #2: At the Foundational Ontological Level of the ontological architecture, in order to comply with the principle of completeness and conciseness along with the principle of delegation of concerns, only one foundational ontology must be found.

• Rule #1: Any new ontology located at level CO, or TDO, or LDO of the ontological architecture depicted in Figure 2 must guarantee a correspondence of its elements with the elements defined at the immediately higher level. For example, to introduce a new ontology at the Core Ontological Level, it must be guaranteed that its elements have a correspondence with the elements defined at the Foundational Ontological Level. This allows the terms and relationships of the lower-level ontologies to be semantically enriched by the terms and relationships of the higher-level ontologies.

• Rule #2: Ontologies of the same level –except at the FO level- can be related to each other, but it must be guaranteed that their joint definition (as a whole) does not violate the principles of the next higher level. This implies that, if a core-level ontology uses elements from another ontology of the same level, together both semantic models must guarantee a correct definition with respect to the foundational-level ontology. This allows the terms and relationships of the ontologies of the same level to complement each other, maintaining a correspondence with the definitions of the ontologies of the higher levels.

• Rule #3: At the Instance Ontological Level, only individuals of particular Things can be found. A Thing like a particular class represented at the foundational level or any of its subclasses (with the semantics of Thing) appropriately represented at the lower levels results in instances. Therefore, an individual is an instance of a particular class at higher levels.

**Acknowledgments.** We warmly thank Maria Julia Blass and Silvio Gonnet (both CONICET researcher and professor at the National University of Technology, Santa Fe, Argentina) for the validation and feedback provided, which allowed us to improve ThingFO from v1.1 to v1.2 and v1.3. Last but not least, I would like to thank Pablo Becker and Guido Tebes (both GIDIS_Web research members at the Engineering School, UNLPam, Argentina) for the close collaboration on the ThingFO specification.

# Appendix A: Updates from ThingFO v1.2 to ThingFO v1.3

Note that ThingFO v1.2 was accepted for publication in an international conference paper [11] held in September 2021, and was published as preprint in April 2021 [8]. The previous versions are ThingFO v1.1 [7], and the first version (v1.0) published in a national conference paper [4], which in turn was first released as preprint in April 2020 [9].

The main updates from ThingFO v1.2 to ThingFO v1.3 are:

- The rephrasing of the terms Thing, Thing Category and Assertion maintaining the same semantics as in the previous versions. The focus was on explicitly adding the word 'class' in the definition of Thing, Thing Category and Assertion, while the word 'expression' was added to the definition of the term Assertion.
- The addition of Notes 4 and 5 to the term Thing, as well as the revision of its synonyms. For example, the former synonym term 'Particular' is now named Particular Thing. Likewise, for the Thing Category main entry, the previous synonym 'Universal' is now Universal Thing.
- The addition of the Note 5 to the term Assertion, as well as the addition of the synonym Human Expression.
- Note 1 for the term 'Thing' was rephrased a bit.
- Note 4 to the term Assertion was rephrased a bit.
- Note 2 for the term 'Relation-related Assertion' was rephrased a bit.
- The removal of three properties of the term Thing Category, namely: 'property category', 'power category' and 'name'. The latter is not necessary since a Thing Category does not result in instances. Therefore, the class name is the name of the category or the names of its subcategories when required. It is also a type since, for example, the term 'Product Category' at the core level [2, 3] has the semantics of Thing Category. On the other side, the two former properties were included as a note in the term (property or attribute) 'descriptive statement'.
- The non-taxonomic relationship 'deals with particulars' was redefined a bit by adding "…both classes/subtypes and instances".
- The Section titled "ThingFO and lower Ontologies in the context of FCD-OntoArch – Guidelines and Rules" is new. Two guidelines and three rules are specified in natural language.
- The ThingFO conceptualization specified in the UML language (Figure 1) has been updated accordingly.
- There are 3 new references ([3], [4] and [8]), while others have been updated a bit.